\documentclass[12pt,reqno]{amsart}
\usepackage{graphicx}
\usepackage{amssymb,amscd,amsbsy}
\usepackage{amsthm}

\usepackage{amsmath}
   \usepackage{amsfonts}   
   \usepackage{amssymb}

\usepackage{graphicx}
\usepackage{amsmath}

\newcommand{\noi}{\noindent}

\def\sign{\text{sign}\,}

\def\W{\mathcal   W}
\def\M{\mathcal   M}
\def\H{\mathcal   H}
\def\E{\mathcal   E}
\def\F{\mathcal   F}

\def\HH{\bold  H}

\def\l{\lambda}
\def\[{\left[}
\def\]{\right]}
\def\({\left(}
\def\){\right)}

\newcommand{\eeq}{\end{equation}}
\newcommand{\beq}{\begin{equation}}
\newcommand{\bay}{\begin{eqnarray}}
\newcommand{\ey}{\end{eqnarray}}
\newcommand{\bey}{\begin{eqnarray*}}
\newcommand{\eey}{\end{eqnarray*}}

\newtheorem{thm}{\hspace{\parindent}Theorem}[section]

\newtheorem{lem}[thm]{\hspace{\parindent}Lemma}
\newtheorem{rem}[thm]{\hspace{\parindent}Remark}

\theoremstyle{remark}
\newtheorem*{rem*}{Remark}

\begin{document}

\newcommand{\vse}{\vspace{.2in}}
\numberwithin{equation}{section}

\title{The AdaBoost Flow}
\author{A. Lykov,  S.Muzychka and K.  Vaninsky}
\begin{abstract}

We  introduce a dynamical system which we call the AdaBoost flow. The flow is  defined by a  system of  ODEs  with control.  
We show that three algorithms of the AdaBoost family (i) the  AdaBoost algorithm of Schapire and Freund (ii) the  arc-gv algorithm  of Breiman  
(iii) the  confidence rated prediction  of Schapire and Singer  can be can be embedded in the AdaBoost flow.  

The  nontrivial part of the AdaBoost flow equations  coincides with the equations of  dynamics of  nonperiodic Toda  system written in terms of spectral variables. We provide a novel invariant geometrical description of the AdaBoost algorithm  as a gradient flow on  a foliation defined by level sets of the  potential function. 

We propose a new  approach for constructing boosting  algorithms as a continuous time gradient flow on measures  defined by various metrics  and potential functions.  
Finally we explain    similarity of the AdaBoost algorithm    with the Perelman's construction for   the  Ricci flow.
\end{abstract}
\maketitle

\tableofcontents
\setcounter{section}{0}
\setcounter{equation}{0}

\section{Introduction.}  The AdaBoost algorithm does not need an advertisement in  data mining community. It was discovered by Robert Schapire and Yoav Freund in their seminal paper in 1997, \cite{FS97}.
Nowadays,   together with the PageRank algorithm, the AdaBoost is  considered  among the top ten algorithm in data mining,  \cite{TT}. It is worth mentioning
that for their AdaBoost paper, Schapire and Freund won the G\"{o}del Prize, which is one
of the most prestigious awards in theoretical computer science, in the year  2003.

The AdaBoost algorithm appeared  from an abstract problem.
In 1988, Kearns and Valiant asked a question whether a  learning
algorithm that performs just slightly better than random guess could be "boosted into an
arbitrarily accurate  learning algorithm. 
Schapire in 1990, \cite{S90},  found that
the answer  is "yes, and the proof he gave is a construction  of the first
boosting algorithm. The AdaBoost proposed  by Freund and Schapire  in 1997, \cite{FS97},  initiated an  extensive research on theoretical aspects of ensemble methods,
which can be easily found in  machine learning and statistical literature. From a practical viewpoint the AdaBoost is 
used to construct spam filtering systems, search engines, face recognition and recommender systems to name a few possibilities.
A Mathematician can forget about all that and treat the AdaBoost as an algorithm which solves some special optimization problem. 

In the present paper we   introduce a dynamical system which we call the \break AdaBoost flow. The flow is  defined by a  system of  ODEs  with control.  
We show that, by a suitable choice of control, three algorithms of the AdaBoost family (i) the  AdaBoost algorithm of Schapire and Freund, \cite{FS97}, (ii) the  arc-gv algorithm  of Breiman , \cite{BR}, (iii) the  confidence rated prediction  of Schapire and Singer, \cite{SS99},  can be can be embedded in the AdaBoost flow. 

The  AdaBoost flow is nothing but  the well known dynamics of   the classical non-periodic Toda  system  of particles on isospectral manifold. 
Introduced in 1967, see \cite{T89},  the Toda lattice  is  a basic 
example of a system of classical mechanics  integrable in the Liouville sense. Its complete integrability  was proved by Moser in 1974, \cite{M74}.  
As an outcome of this  connection  we provide a novel geometrical description of the AdaBoost as a  gradient flow on a foliation defined by level sets of a potential function. 

We propose a general approach for constructing AdaBoost like algorithms  using gradient flows defined by various metrics  and potential functions. We   also introduce  a new continuous time algorithm which we call SuperBoost.
Finally we discuss   similarity of the AdaBoost algorithm  with the Perelman's approach  to control the  Ricci flow.  We show that all parts of the  Schapire and Freund construction   have their counterparts in the Perelman's construction. We present  a dictionary between two problems.    It turns out, the AdaBoost and the Perelman's construction  are different realization of the same idea. 

The idea can be described vaguely as follows. Suppose we have a space $X$ which is mapped into a space $Y$ by a  family of maps  $f_t,$ where $t\in \[ 0, +\infty\)$. 
The space of all such maps $f_t$  is denoted by $\F$. We are interested in the behavior of maps when $t$ becomes large. For example,  $X$ can be a manifold and $Y$ can be a space of symmetric nonnegative matrices. In this case $f_t$   define a metric on $X$ for each $t$. The family $f_t$ can be produced by  the  Ricci flow. 
 
Suppose that there are points $x_1,..., x_N;$ where the maps $f_t$ behave badly  when $t$ becomes large, but we still want to control the maps around these points.   In order to do this  an extension of the original dynamics on the  space  $\F$ is introduced. Let $\W$ be a space of all  probability measures on $X$. 
At  each moment $t$ the probability measure $w_t$ is chosen such that some (integral)  functional $I$  of two variables increases with time
$$
\frac{d}{dt} I(f_t,w_t) \geq 0.
$$ 
In other words  the functional $I$ plays the role of a Lyapunov function for the dynamics on the extended phase space $\F\times \W$. Of course,  there are  different variations  of the described construction. For example $w_t$ can be replaced by $w_0$ as in the case of the AdaBoost.  The point of this construction is that monotonicity of the functional $I$ allows to control the maps  $f_t$  around "bad" points.

The paper consists of two sections. In the first section we present all necessary facts about classical discrete AdaBoost algorithm.   
For many of these facts we establish continuous time counterparts.  In the second section we introduce continuous 
time AdaBoost flow and consider embeddings of three algorithms of the AdaBoost family  
into  continuous time flow. We demonstrate  gradient character of the AdaBoost flow  and  introduce a general approach to boosting based on 
continuous time gradient flows. Finally we show  that the AdaBoost  and the Perelman's construction  for  the Ricci flow are closely related.

We would like to thank Igor Krichever, Sasha Veselov and Vadim Malyshev for stimulating discussions.  Our special thanks to Henry McKean for  numerous remarks.

\newpage
\section{Discrete time  AdaBoost algorithm.}
\subsection{Basic AdaBoost  algorithm. } We start with    simple description of the AdaBoost algorithm.  The reader can consult 
the book \cite{FS12} for an additional information.

Given   a  set of points $$TS=\{(x_1,y_1),..,(x_m,y_m)\},$$  where $x_i \in X$ and $y_i \in \{-1,+1\}$.   Usually $X=R^d$ and $y_i$ are called labels.  Such set $TS$ is called a training set and it will be used to construct a decision rule. Also given  a finite set of  functions    $\H_0=\{ h_{\gamma},\; \gamma \in \Gamma\}$; where  each
$h_{\gamma} \in \H_0$ is such that  $h_{\gamma}: X \rightarrow \{-1,+1\}.$ These functions are  called weak classifiers. Pick some probability distribution $w(i),\;  i=1,..., m;$ on the  training set.   The classification error of  any weak  classifier $h_{\gamma}$  with respect to the measure $w$ is defined as 
$$
W^-(h_{\gamma},w)=  w  \{i: h_{\gamma}(x_i) y_i=-1\}. 
$$
In a practical situation this error can be quite big, {\it i.e. } $W^-(h_{\gamma},w) >>0$.  That is the reason to call $h_{\gamma}$  weak classifier.  To remedy  large classification error one can try to combine  weak classifiers.    

Let  $\H$ be a positive cone over a set of basic classifiers
$$
\H=\{H: H=\sum_{\gamma \in \Gamma} \alpha_{\gamma} h_{\gamma};\,  h_{\gamma} \in \H_0,\;   \alpha_{\gamma} \geq 0\}.
$$
From any $H$,  the combined classifier $\HH: X \rightarrow \{-1,0,+1\}$ can be  constructed.  Namely, if $H(x)\neq 0 $, then  $\HH(x)=\sign H(x)$;
if   $H(x)= 0 $, then  no decision can be made and $\HH(x)=0$.
Let us define
$$
W^-(H,w) =  w  \{i: \HH(x_i) y_i=-1 \}\qquad\qquad W^0(H,w) =  w  \{i: \HH(x_i) y_i=0 \}. 
$$
The problem   is to minimize the error $W^-+ W^0$ of   the   combined classifier $\HH$ 
by choosing appropriate values of $\alpha_{\gamma}$. The  difficulty of  this constrained minimization problem is that the error   is almost everywhere constant on $\H$ and
gradient methods can not be applied  directly.

The AdaBoost algorithm offers a candidate for the  solution of  this problem in a series of $N+1$ rounds,   where $N$ is   some integer number.
For any $n=0,1,...,N;$ the combined  classifier
$\HH_n(x)=\HH_n: X \rightarrow \{-1,0,+1\}$ is defined as
$$
H_n= t_0 h_{\gamma_0}+ ... + t_n h_{\gamma_n};
$$
where the sequence of positive  numbers  $t_0, t_1, ..., t_n,$ is constructed simultaneously with   $h$'s. The final classifier $\HH=\HH_N$ is a candidate for the solution of the minimization problem.  In practical applications one   simply chooses $N$  large enough. Theoretical bound for the misclassification error  will be given below.

Specifically, the  AdaBoost  recursively constructs a family of classifiers by means of  probability measures $w_0, w_1, ..., w_{N}$. It  starts with the fixed  distribution $w$:
$$
w_0:\quad w_0(i)= w(i), \qquad i=1, ..., m.
$$
Given a distribution $w_n, \; n=0,..., N;$ the  AdaBoost algorithm picks arbitrary  $h_{\gamma_n}$  from $\H_0$ such that
\beq\label{kkk}
W_n^-=    W^- (h_{\gamma_n},w_n)< 1/2.
\eeq
If  at  some step it is not possible to do this, {\it i.e.} if 
$$
 \min_{h_{\gamma} \in \H_0}  W^- (h_{\gamma},w_n) \geq 1/2;
$$
then the procedure stops unfinished.  The reason for this will be explained later. Note that on each step  the algorithm   does not have to go through the whole  set $\H_0$, one has 
just  to find one $h_\gamma$  that satisfies \ref{kkk}.
If this is the case and $W^-_n < 1/2 $ for all $n=0,1,..., N;$
then the measure is constructed recurrently
$$
w_{n+1}(i)= \frac {e^{- t_n y_i h_{\gamma_n}(x_i)}w_{n}(i) } {Z_n},
$$
where  $t_n$ is some positive number and
$$
\qquad Z_n=  \sum_{i=1}^{m} e^{- t_n y_i h_{\gamma_n}(x_i)}w_{n}(i) .
$$
The whole procedure can be represented by a diagram
$$
\setlength\arraycolsep{0.1em}
 \begin{array}{rclclcl}
  H_0         & \phantom{h_0}&        H_1     & \phantom{h_0}&   ...    &   \phantom{h_0}&  H_N\\
  \uparrow     & \searrow&             \uparrow & \searrow&       ...     &  \searrow& \uparrow \\
   w_0        & \phantom{h_0}&        w_1         & \phantom{h_0}&  ...   &    \phantom{h_0}&  w_N
 \end{array}
$$
The function $H(x)=H_N(x)$ takes values in the segment $\[-T,+T\]$, where $T =\sum_{n=0}^{N} t_n$.

At each step of the AdaBoost procedure the set of training points $TS$ falls into two categories  $G_n$  (good)  and $B_n$ (bad). 
Points of $G_n$ are  classified correctly by $h_{\gamma_n}$
$$
G_n=\{(x_i,y_i): h_{\gamma_n}(x_i)y_i=+1\}.
$$
The measure $W^+_n=w_n\{G_n\}$ of these points  decreases upon the next step
$$
w_n(i) \rightarrow w_{n+1}(i) =\frac{e^{-t_n}}{Z_n}w_n(i), \qquad \qquad\qquad\qquad  (x_i,y_i) \in G_n.
$$
The points of $B_n$ are   misclassified by $h_{\gamma_n}$:
$$
B_n=\{(x_i,y_i): h_{\gamma_n}(x_i) y_i=-1\}.
$$
The measure $W^-_n=w_n\{B_n\}$ of these points  increases upon the next step:
$$
w_n(i) \rightarrow w_{n+1}(i) =\frac{e^{t_n}}{Z_n}w_n(i), \qquad \qquad\qquad\qquad  (x_i,y_i) \in B_n.
$$
Apparently, $W^+_n + W^-_n=1$ and
$
W^-_n < \frac{1}{2}, \; n=0,1,...,N.
$
The values of $t_n$ at each step are chosen to minimize probability  of error of the final combined classifier.
Remark \ref{int} shows that with an optimal choice of $t_n$  
$$
w_{n+1}\{ G_n\}=w_{n+1}\{ B_n\}= \frac{1}{2}.
$$



\subsection{AdaBoost map on the extended phase space}
Boosting can be viewed as a discrete time dynamical system on the extended phase space  $\H\times \W$  which is the  direct product  of the positive cone $\H$ and  
the simplex  of probability measures $\W=\{w: \sum_{i=1}^{m} w(i)=1,\; w(i)\geq 0\}$. The  vector field $v(H,w)$ on $\H\times \W$  is a constant $h=h(w)$ on the  "fibers"
$\H_w=\{(H,w'):\; H\in \H,\; w'=w\}$, {\it i.e.}
$$
v(H,w)=h,\qquad\qquad \text{for any}\;  H\in \H_w.
$$
The AdaBoost dynamics maps $(H_{n}, w_{n})$ into $(H_{n+1}, w_{n+1})$ by the rule
\beq
w_{n+1}(i)= \frac {e^{- t_n y_i  v(H_{n}, w_{n})  (x_i)} w_n(i)} {Z_n},\qquad \qquad n=0,1,...;\;  i=1,2, ..., m;
\eeq
\beq
H_{n+1}=H_{n}+ t_{n+1} v(H_{n}, w_{n+1}),\quad\quad n=-1,0,1,....
\eeq

Along the trajectory  $\{(H_n, w_n), \; n=0,1,...,N\}$  the  Adaboost dynamics  drives the error 
$$W^-(H_n, w_0)+ W^0(H_n, w_0)=\sum_{i=1}^{m} w_0(i) \chi_{[y_i H_n(x_i)\leq 0]} (x_i)
$$  
to zero. To prove this one  overestimates  the function $W^-(\cdot, w_0)+ W^0(\cdot, w_0)$ which has no   gradient by  some smooth function $\E(\cdot,w_0)$ defined as
$$
\E(H,w)= \sum_{i=1}^{m} w(i) e^{-y_i H(x_i)}.
$$
Clearly,
$$
W^-(H, w_0)+ W^0(H, w_0)=\sum_{i=1}^{m} w(i) \chi_{[y_i H(x_i)\leq 0]} (x_i) \leq \E(H,w)= \sum_{i=1}^{m} w(i) e^{-y_i H(x_i)}.
$$
The  function $\E(\cdot, w_0)$ plays the role of Lyapunov function  for the AdaBoost dynamics.
It is strictly convex in the first  argument
$$
\E(\lambda H' +\mu H'',w)< \lambda\, \E(H',w) + \mu\, \E(H'',w),
$$
and linear in the second.  

Equations of the Adaboost dynamics imply  that the  values of $\E(H_n, w_0)$ along the trajectory   satisfy two equivalent identities. The first connects two consecutive values
\beq\label{local}
\E(H_{n+1},w_0)= Z_{n+1} \E(H_{n},w_0).
\eeq
The second identity  reads
\beq\label{glob}
\E(H_{n},w_0)=\prod_{p=0}^{n}  Z_p.
\eeq
The  identities   can be easily proved by means of  the relation
\beq\label{impid}
\E(H_n, w_k)= Z_k\times \cdots \times Z_n\,  \E(H_{k-1},w_{n+1}),\qquad\qquad  k< n;
\eeq
and boundary condition $\E(H_{-1},w_n)=1$, due to  $H_{-1}=0$. Both identities have their counterparts in the continuous time case.

The constant $t_n > 0$ is chosen  to minimize $Z_n$ on each step. In detail,
$$
Z_n(t)=  e^{-t} W^+_n + e^{t} W^-_n,
$$
and  from the  condition of critical point $\frac{d Z_n}{d t}=0 $  we get
an explicit   formula
$$
t_n= \frac{1}{2} \log \frac{W_n^+}{W_n^-}.
$$
The constant $t_n$ is positive if  and only if $W^-_n < 1/2$.
The formula  for    $Z_n$
$$Z_n= 2 \sqrt{ W^+_n W^-_n };$$
with optimal $t_n$ follows  easily.
If $W^-_n= \frac{1}{2} - \beta_n$,  then
$$
Z_n =   \sqrt{ 1- 4\beta_n^2}\leq  e^{-2 \beta_n^2},
$$
and
$$
W^-(H_N, w_0)+ W^0(H_N, w_0)  \leq \E(H_N,w_0)\leq e^{-2 \sum_{p=0}^{N} \beta_p^2}.
$$
Therefore  the training error decays exponentially with $N$,  if $  \beta_n$ are uniformly bounded from zero. Moreover, if $N$ is such that 
$$
\min_{i} w_0(i) > e^{-2 \sum_{p=0}^{N} \beta_p^2}, 
$$
then $  W^-(H_N, w_0)+ W^0(H_N, w_0)  =0 $.

\begin{rem}\label{int}
Note that formulas for $Z_n$ and $t_n$  imply
$$
 w_{n+1}\{B_n\}= \frac{e^{t_n}}{Z_{n}} \, W^-_n = \frac{e^{t_n} W^-_n }{2 \sqrt{ W^-_n W^+_n}}= \frac{1}{2}.
$$
\end{rem}

\subsection{AdaBoost as entropy projection.} 
The fact that updates of the measure $w_n$ on each step are solutions of an extremal problem is due to Kivinen and Warmouth, \cite{KW}. 
It can be checked directly that $w_{n+1}$ is a solution of the following minimization problem.

Given $w_n$ find such $w$ that it minimizes Kulback-Leibler divergence 
$$
KL(w|| w_n)=\sum_{p=1}^{m} w(p)  \log \frac{w(p)}{w_n(p)}, 
$$ 
subject to constraints 
\beq\label{hyp}
\sum_i y_i h_n(x_i) w(i)=0,\qquad
\eeq
and 
$$
\sum_i w(i)=1.\qquad
$$
The solution of this problem $w_{n+1}$ is given  by a projection of $w_n$ into a hyperplane defined by  condition \ref{hyp}. 
Usually  such a projection is given by a geodesic  curve  which is a solution of the second order equation.

\newpage
\section{Continuous  time  AdaBoost flow.}
The first observation   of our paper can be described as follows.  We will write a gradient flow which is  a system of  the first order equations. 
The solution of these equations is a continuous curve which connects $w_n$ 
and $w_{n+1}$.  The time it takes for the trajectory of the gradient flow to travel from $w_n$ to $w_{n+1}$ is exactly the weight $t_n$.  
In order to write such a  gradient flow we  use the metrics which arises from the   KL divergence and a potential function 
associated with $h_n$. 

\subsection{Differential equations for the AdaBoost flow.}  In this section we introduce a  continuous time AdaBoost flow on the  $\H \times W$. Namely  we construct a family of combined classificators  $H_t$  and  measures  $w_t$,  for all $t,\; 0 \leq t \leq T.$  Naturally, 
$$
H(x_1), H(x_2),..., H(x_m), \qquad\quad w(1), w(2),..., w(m);
$$
are coordinates on $\H \times W.$

Let $\gamma_t: [0,\infty)  \rightarrow \Gamma$ be  a function with finite number of values on any finite interval  and let it be    continuous from the right  
with respect to the  discrete topology on $\Gamma$. We choose  a vector field  constant on   $\H_{w_t}$ as in discrete case, {\it i.e.}
$$
v(H_t,w_t)=h_{\gamma_t}. 
$$
In this language  the AdaBoost flow differential equations are the following
\beq\label{evcl}
 \frac{d}{d t}\; H(x_k) = \quad   v(H_t,w_t)(x_k) ,  \qquad\qquad \qquad\qquad \qquad   k=1,2, ..., m;
\eeq
\beq\label{evme}
\frac{d}{d t}\;  w_t{(k)}= -   y_k v(H_t,w_t)(x_k) w_t(k) +   \sigma_t w_t(k) , \qquad k=1,2, ..., m;
\eeq
where  $\sigma_t=\sigma_{w_t}=\sum_{p=1}^m y_p v(H_t,w_t)(x_p)  w_t(p).$ It can be checked easily that the  quantity   
$$
w(1)+w(2)+ ... + w(m)
$$
is an integral of motion. Therefore, the orbits  of the AdaBoost flow remain   on the simplex $W$ for all times.

Differential equation allows us to define the AdaBoost flow when  the  weak classifiers take    arbitrary real values, {\it i.e.}    we assume that 
$h_{\gamma}: X \rightarrow R^1;$ for any  $h_{\gamma} \in \H_0.$  
The solution of the differential equations with a fixed  $\gamma_t$ is a  straight line motion 
$$
H_t=H_0+ t\times v(H_0,w_t) ; 
$$
and   the measure   is  
$$
w_t(k)= \frac{w_0(k) e^{-t y_k v (H_0, w_0)(x_k)}}{\sum_{p=1}^{m} w_0(p) e^{-t y_p v (H_0, w_0)(x_p)}}, \qquad\qquad \qquad k=1,2, ..., m.
$$

The equations for $w_t(k),\; k=1,..., m;$ coincide with the equations for  spectral weights in \cite{M74} and \cite{V03}. 
In the case of Toda lattice all the numbers $y_k v(H_t,w_t)(x_k)$ are distinct  for different $k$. They are the simple spectrum of the Jacobi matrix.
Here the situation is different.
In the case of  weak classifiers which take only two values $+1$ and $-1$, the components $y_k v(H_t,w_t)(x_k)$ of the vector field  also  take
only these two possible values.

Now we want to write the equations   \ref{evme}  in a different form. 
This new form    will be used in section 3.7. 
The measure $w$ can be defined in terms of the potential function $f$ as   $w_t(k)= e^{- f_t(k)},\; k=1,..., m$.  Then we can write  \ref{evme} as 
\beq\label{evmepot}
 \frac{d}{d t}\;  f_t{(k)}=    y_k v(H_t,w_t)(x_k)  -   \sigma_t  , \qquad k=1,2, ..., m;
\eeq

What are the orbits of the AdaBoost flow  on the simplex  $\W$?
Let weak classifiers  take values  $\{+1,-1\}$. Define
$$
W^{+}=w_{0}\{i:y_{i}h(x_{i})=1\},$$
$$
W^{-}=w_{0}\{i:y_{i}h(x_{i})=-1\}>0;
$$
and
$$
U(t)=\frac{1}{W^{+}+e^{2t}W^{-}},\qquad t\geqslant0.
$$
\begin{lem}
Assume that the AdaBoost flow runs for all $t \geq 0$  with
the fixed $v(H_{t},w_{t})$ i.e. it comes from a fixed single  classifier $h$.
Then for 
$$
w_{t}=\mathcal{L}[w_{0}]+\mathcal{D}[w_{0}]U(t),$$
where the vectors \textup{$\mathcal{L}[w_{0}]$} and $\mathcal{D}[w_{0}]$
are defined by the following formulas: 
$$
\mathcal{L}[w_{0}](i)=\begin{cases}
0 & y_{i}h(x_{i})=1\\
\frac{w_{0}(i)}{W^{-}} & y_{i}h(x_{i})=-1\end{cases}\quad \qquad\qquad i=1,\ldots,m;$$
$$
\mathcal{D}[w_{0}](i)=\begin{cases}
w_{0}(i) & y_{i}h(x_{i})=1\\
-\frac{W^{+}}{W^{-}}w_{0}(i) & y_{i}h(x_{i})=-1\end{cases}\qquad\qquad i=1 ,\ldots,m;$$
\end{lem}

{\it Proof.}
Substitute  explicit expression for the flow into the formulas.
\qed

Since for  $t\geqslant0$,   $U(t)\in(0,1]$,  the orbit of the point
$w_{0}$ under the AdaBoost flow is a semi-interval between the points   $w_{0}$
and  $\mathcal{L}(w_{0}$). Moreover, as  $t\rightarrow+\infty$,
$w_{t}\rightarrow\mathcal{L}[w_{0}]$, {\it i.e.} the AdaBoost flow transports the measure towards the points where the classificator makes an error.

Now let  weak classifiers  take values in $\{-1,0,+1\}$.
Define:
$$
W^{0}=w_{0}\{i:h(x_{i})=0\}.
$$
Let us assume  $0<W^{0}<1$.
Define:
$$
Z(t)=W^{+}e^{-t}+W^{-}e^{t}+W^{0}
$$
$$
\alpha(t)=\frac{e^{-t}}{Z(t)},
$$
$$
\beta(t)=\frac{1}{Z(t)}.
$$

\begin{lem}
For any  $ t\geqslant0$, 
$$
w_{t}=\mathcal{L}[w_{0}]+\mathcal{D}^{+}[w_{0}]\alpha(t)+\mathcal{D}^{0}[w_{0}]\beta(t),
$$
where the vectors \textup{$\mathcal{D}^{+}[w_{0}]$} and $\mathcal{D}^{0}[w_{0}]$
are defined as:
$$
\mathcal{L}[w_{0}](i)=\begin{cases}
\frac{w_{0}(i)}{W^{-}} & y_{i}h(x_{i})=-1\\
0 & \text{otherwise} \\
\end{cases}\quad \qquad\qquad i=1,\ldots,m;
$$

$$
\mathcal{D}^{+}[w_{0}](i)=\begin{cases}
w_{0}(i) & y_{i}h(x_{i})=1\\
0 & h(x_{i})=0\\
-\frac{W^{+}}{W^{-}}w_{0}(i) & y_{i}h(x_{i})=-1\end{cases}\quad i=1,\ldots,m;
$$
$$
\mathcal{D}^{0}[w_{0}](i)=\begin{cases}
0 & y_{i}h(x_{i})=1\\
w_{0}(i) & h(x_{i})=0\\
-\frac{W^{0}}{W^{-}}w_{0}(i) & y_{i}h(x_{i})=-1\end{cases}\quad i=1,\ldots,m.
$$
Moreover,  functions  $\alpha$ and  $\beta$ satisfy the equation
$$
a\alpha^{2}+d\alpha\beta+b\beta^{2}-\alpha=0, 
$$
where $a=W^+,\; b=W^-$  and $d=W^{0}$. 
\end{lem}
{\it Proof.}
The equations  follow from the obvious relations
$$
\frac{\alpha}{\beta^{2}}=Ze^{-t},
$$
$$
\alpha-\frac{1}{a}=-\frac{b}{a}\frac{1}{Ze^{-t}}-\frac{d}{a}\frac{1}{Z}.
$$
\qed

As in the first case  when classifier takes only two values, $w_{t}\rightarrow\mathcal{L}[w_{0}]$ when 
$t\rightarrow+\infty$. In the present  case with three values,  the orbit of $w_0$ lies in a two dimensional plane on a second degree algebraic curve.

\begin{lem} Assume that the AdaBoost flow runs on the infinite time interval with the same $v(H_t, w_t)$ {\it i.e.} it comes from the same classificator.
Let
$$
V_{\max}= \max_k y_k v(H_t,w_t)(x_k) \qquad\quad {\text and} \quad\qquad V_{\min}= \min_k y_k v(H_t,w_t)(x_k).
$$
Then, $\frac{d}{dt} \sigma(t) <0$ and $\lim_{t\rightarrow -\infty} \sigma(t)=V_{\max},\quad \lim_{t\rightarrow +\infty} \sigma(t)=V_{\min}$.

\end{lem}

{\it Proof.}          By Jensen's  inequality and the strict convexity of the function $x^2$,  we get
\bey
\frac{d}{dt}\,  \sigma(t)&=&\sum_{p=1}^m y_p v(H_t,w_t)(x_p)  \[ -   y_p v(H_t,w_t)(x_p) w_t(p) +   \sigma_t w_t(p)   \]=\\
&=&- \sum_{p=1}^m \[y_p v(H_t,w_t)(x_p)\]^2  w_t(p) +   \sigma_t^2  <0;
\eey
The rest can be proved easily.  \qed

The next result for the Lyapunov function is central for our discussion. This identity is a continuum  analog of \ref{impid}.

\begin{thm}  If  $p<t,$ then
 $$
 \log \E(H_t,w_p)-  \log \E(H_p,w_t) = -\int_{p}^{t}   \sigma_s \,  ds.
$$
\end{thm}

{\it Proof.} Assume that the AdaBoost flow runs on the  time interval $\[ p, t \]$ with the same $v(H_s, w_s)$ {\it i.e.} it comes from the same classificator $h_{\gamma}$.
In general, the interval $\[ p, t \]$ can be split into  subintervals with this property.
Differential equations imply two identities
$$
H_t= H_p + \int_{p}^{t}    v(H_s,w_s) \, ds;
$$
on such subintervals,  and for any $k=1,..., m;$
$$
w_t(k)= w_p(k) \, e^{-\int_p^t   y_k v(H_s, w_s) (x_k) ds  } \,  e^{\int_p^t    \sigma_s ds  } .
$$
Therefore,
$$
\E(H_t,w_p)= \sum_k w_p(k) e^{-y_k H_t(x_k)}= \sum_k w_p(k) e^{-\int_p^t  y_k v(H_s,w_s) (x_k) ds} e^{-y_k H_p(x_k)}=
$$
$$
= \sum_k w_t(k) e^{-\int_p^t  \sigma_s ds} e^{-y_k H_p(x_k)}=  \E(H_p,w_t)\;  e^{ -\int_{p}^{t}   \sigma_s \,  ds}.
$$
\qed

Using the fact that,  $\E(H_0,w_p)=1$,  we obtain the analog of \ref{local}
\beq\label{AB}
\frac{d}{ d t} \log \E(H_t,w_0) = - \sigma_t,
\eeq
and the  analog of \ref{glob}
\beq\label{cglob}
 \E(H_T,w_0)  =  e^{- \int_0^T    \sigma_s \; ds}.
\eeq
The last identity implies that one should try to choose  $\gamma_t$ so that $\sigma_t$  is maximal along the path. In fact, there are a few  choices. 
As it will be shown below, 
they correspond to the discrete AdaBoost algorithms, the arc-gv algorithm and the AdaBoost with varying confidence level.

\subsection{Embedding of the discrete AdaBoost  into the AdaBoost flow.}

In this section we assume   that  all weak classifiers $h_{\gamma},\; \gamma \in \Gamma;$ take only two values $-c$ and $+c,\;c> 0. $ The formulas obtained in this section will be generalized for the case of classificators with varying confidence level.  For each classificator,  we  define
$$
W^-=w_0\{i: y_i h_{\gamma}(x_i)=-c \}\qquad\qquad W^+=w_0\{i: y_i h_{\gamma}(x_i)=+c\}.
$$
We also  assume that $ 1/2 <   W^+ < 1.$ 

\begin{thm}\label{pop} Let the AdaBoost flow runs up to time  $\Delta$ with fixed  $v(H_t, w_t)$ {\it i.e.} it comes from a fixed  classificator  $h_{\gamma}$. Then, 

(i) For any $\Delta >0$, 
\beq\label{tt}
 e^{ -\int_{0}^{\Delta}   \sigma_s \,  ds} \geq  2\sqrt {W^+  W^-}.
\eeq

(ii) The equality in \ref{tt} holds if and only if
\beq\label{opt}
\Delta= \frac{1}{2 c} \log \frac{W^+}{W^-}.
\eeq

(iii) The equality in \ref{tt} holds for some $\Delta>0 $ if and only if $\sigma_{\Delta}=0.$
\end{thm}

{\it Proof.} (i.) By \ref{cglob},   
$$
e^{-\int_0^\Delta \sigma_s \, ds}= \sum_{k=1}^m  w_0(k) e^{-\Delta y_k v(x_k)} =W^+e^{-\Delta c}+ W^{-} e^{\Delta c}.
$$
Inequality \ref{tt} follows from the inequality between the arithmetic  and geometric means.

(ii.) Write, 
$$
Z(\Delta,c)= e^{-\int_0^\Delta \sigma_s \, ds}.
$$
If  left hand side  of  \ref{tt} attains its minimum and equality holds,  then
$$
\frac{\partial Z}{\partial \Delta}=0.
$$
This implies the formula  \ref{opt}. The converse statement can be checked directly.

(iii.) The condition $\sigma_{\Delta}=0$ means  that 
$$
\sigma_{\Delta}=W^-e^{\Delta c}- W^{+} e^{-\Delta c}=0,
$$
This is equivalent to  \ref{opt}  and so  to  \ref{tt}  as well.
\qed

To explain the connection  between continuous time system and the AdaBoost algorithm  we  assume that $c=1$.
One can define the values of the control  $\gamma_t: \[0,+\infty\) \rightarrow \Gamma,$ recurrently by the following procedure. 
Given $w_0$ and  for $t_{-1}=0$, define 
\beq\label{min}
\gamma_0=\gamma_{t_{-1}}= \arg \min_{\gamma \in \Gamma} W^-(h_\gamma, w_0).
\eeq
Supposed that   $W^-=W^-(h_{\gamma_{0}}, w_0)  < 1/2$. Then  $\sigma_0=W^+-W^- >0$  decays with time. The AdaBoost flow runs with this $\gamma_0$ until the   time  $t_0=\Delta$; that can be computed   from  \ref{opt}. Note that $\sigma_\Delta=0.$
Therefore, we define
$
\gamma_t= \gamma_{0}
$
for $t\in [0,t_0)$. It is interesting to check that $$w_{t_0}\{i: y_i h_{\gamma_0} (x_i) < 0\} =\frac{1}{2}.$$
For  the next step one should look for a new solution of the minimization problem \ref{min} with $w_0$ replaced by $w_{t_0}$, {\it etc.}

In general,  put
$
t_n=\sum_{p=0}^{n} \Delta_p,
$
for $n\geq 1$. The corresponding control and errors  are 
$$
\gamma_{t_{n-1}}= \arg \min_{\gamma \in \Gamma} W^-(h_\gamma, w_{t_{n-1}}),
$$
and $W^-_n= W^-(h_{\gamma_{t_{n-1}}}, w_{t_{n-1}})  < 1/2$.
The intervals $\Delta_n$ are determined from  \ref{opt}:
$$
\Delta_{n}=\frac{1}{2c} \log\frac{W_n^+}{W_n^-}= \frac{1}{2} \log\frac{1+2\beta_{t_{n-1}}}{1- 2\beta_{t_{n-1}}},
$$
where
$$
W_n^-= W^-(h_{\gamma_{t_{n-1}}}, w_{t_{n-1}})=\frac{1}{2} -\beta_{t_{n-1}}.
$$
It is easy to see that 
$$
w_{t_{n}} \{i: h_{\gamma_{t_{n-1}}}(x_i) y_i=-1\}=\frac{1}{2}.
$$
The sequence of $(H_{n}, w_n)= (H_{t_n}, w_{t_n})$ is also a trajectory of the discrete AdaBoost algorithm.




\subsection{Embedding of the arc-gv algorithm into the   AdaBoost flow.}

First we formulate a version of the discrete algorithm, see \cite{BR}. 

Assume that we have a classifier 
$$
H=\sum_{k=0}^{n}t_{k}h_{k},
$$
where  $t_{k}>0$ and $h_{k}\in\mathcal{H}_0$,  for  $k=0,\ldots,n$. Introduce the  norm,
$$
\|H\|=\sum_{k=0}^{n}t_{k};
$$
together with the normalized margin of the function $H$ at the point $(x,y)\in X\times\{-1,+1\}$
$$
m(x,y;H) =  y\frac{H(x)}{\|H\|},
$$
and the minimal margin
$$
\mu(H)  = \min_{(x,y)\in TS}\{m(x,y;H)\}.
$$
Let  $\mu(0)=-1$ and note the obvious properties of $\mu(H)$. First,  $-1\leqslant\mu(H)\leqslant1.$
Second, $\mu(H)=-1$ if and only if there exists $(x,y)\in TS$ such that $h_{k}(x)\ne y$  for all $k=0,\dots,n$, {\it i.e.}
there is a point that all weak classifiers constituting $H$, make on error or  $H=0$.
Third, assume, that $\mu(H)=1$, then for all $(x,y)\in TS$ one has  $h_{k}(x)=y$, where $ k=0,\ldots,n;$
in other words all weak classifiers are  able to separate points without error.
We assume that there are no such classifiers at all  and   $\mu(H)<1$.  Fourth, $\mu(H)>0$ if and only if for all $(x,y)\in TS$ 
one has $ yH(x)>0$,  {\it i.e.} all points are classified correctly by the function  $H$.

\noi 
Now we describe the arc-gv algorithm itself. 

Initialization,
\begin{itemize}
\item  $H_{-1}=0,$
\item  $w_{0}(i)=\frac{1}{m},\; i=1,\ldots,m,$
\item  $\widetilde{t}\gg1$ - regularization parameter (large positive number).
\item  For $n=0,\dots$
\end{itemize}

For $n=0,1,...$,
\begin{itemize}
\item Choose a weak classifier  $h_{\gamma_{n}}\in\mathcal{H}_{0}$:
$W^{-}(h_{\gamma_{n}},w_{n})<\frac{1}{2}$;
\item $\beta_{n}=\frac{1}{2}-W^{-}(h_{\gamma_{n}},w_{n});$
\item $\mu_{n-1}=\mu(H_{n-1});$
\item Determine the weight: $t_{n}=\min\{\tilde{t},\frac{1}{2}\ln(\frac{1+2\beta_{n}}{1-2\beta_{n}})-\frac{1}{2}\ln(\frac{1+\mu_{n-1}}{1-\mu_{n-1}})\}$;
\item If $t_{n}\leqslant0$, then the algorithm stops;
\item Update the measure: $w_{n+1}(i)=\frac{1}{Z_{n}}\exp(-t_{n}y_{i}h_{\gamma_{n}}(x_{i}))w_{n}(i)$;
\item $H_{n}=H_{n-1}+t_{n}h_{\gamma_{n}}$.
\end{itemize}

\noi
The formula for the weight  $t_{n}$ appears, see \cite{BR},  from the following optimization problem:  
minimize in  $t\in[0;\tilde{t}]$ the function
$$
\Theta(t)=\sum_{i=1}^{m}w_{n}(i)e^{t(-y_{i}h_{\gamma_{n}}(x_{i})+\mu_{n-1})}.
$$
As for the  AdaBoost,  one finds an exact formula for the optimal $t$  by differentiation.
Moreover, for  $\mu_{n-1}\ne\pm1$ :
$$
Z_{n}=\sqrt{W_{n}^{-}W_{n}^{+}}\(\sqrt{\frac{1-\mu_{n-1}}{1+\mu_{n-1}}}+\sqrt{\frac{1+\mu_{n-1}}{1-\mu_{n-1}}}\),
$$
$$
w_{n+1}\{i:h_{\gamma_{n}}(x_{i})\ne y_{i}\}=\frac{1-\mu_{n-1}}{2}.
$$
The embedding  of arc-gv into  AdaBoost flow is the same as for the discrete AdaBoost algorithm.  Note that:
$$
\mu_t=\mu(H_{t})=\frac{1}{t}\min_{(x,y)\in TS}\{yH_{t}(x)\}.
$$
The formulas for embedding are similar:
$$
H_0=0;
$$
$$
\Delta'_{n}=\min\{\overline{\Delta},\frac{1}{2}\ln\(\frac{1+2\beta_{t_{n-1}}}{1-2\beta_{t_{n-1}}}\)-\frac{1}{2}\ln\(\frac{1+\mu_{t_{n-1}}}{1-\mu_{t_{n-1}}}\)\},\quad n\geqslant 0;
$$
where  $\overline{\Delta}$ is a large fixed number. If at some moment $\Delta'_{n}\leqslant0$, then the algorithm stops.

The general picture is as follows: At the beginning, when
$\mu_{t_{n}}=-1$, we switch classifiers at the equal intervals $\overline{\Delta}$.
Then  $\mu_{t}>-1+\epsilon$
and the algorithm starts to switch at  smaller intervals than $\overline{\Delta}$, but bigger then  prescribed by  AdaBoost. That happens until  $\mu_{t}\leqslant0$.
At some moment  $\mu_{t}=0$   which is such that  constructed classifier $H_{t}$ learned how to separate points without error. Finally,  as a protection from overfitting,  the algorithms stops when $\mu_{t_{n}}>2\beta_{t_{n}}$.

\subsection{Embedding of the CRP  algorithm   into the AdaBoost flow.}  In this section we will show how confidence rated prediction (CRP)  of Schapire and Singer, \cite{SS99},  
can be embedded into the AdaBoost flow. 
Let the set of all values of $h_{\gamma}$ be 
$c_j, \; j=1,..., p;$ and take 
$$
W^{+,j}=w_0\{i: \;  h_{\gamma}(x_i)= c_j,\; y_i=+1  \},
$$
and
$$
W^{-,j}=w_0\{i: \;  h_{\gamma}(x_i)= c_j,\; y_i=-1  \}.
$$

\begin{thm} Fix some   $\Delta>0$. Let the AdaBoost flow runs up to    time  $t= \Delta$ with  fixed  $v(H_s, w_s)$ {\it i.e.} it comes from the fixed classifier $h_{\gamma}$.

(i) Let  $W^{+,j}\, W^{-,j} >0$ for all $j=1,..., p;$ then for any  $c_j$  
\beq\label{ttt}
 e^{ -\int_{0}^{\Delta}   \sigma_s \,  ds} \geq  \sum_{j=1}^{p} 2\sqrt {W^{+,j}\;  W^{-,j}}.
\eeq

(ii)    Let   $W^{+,j}\, W^{-,j} >0$ for all $j=1,..., p;$ then  the equality  holds in \ref{ttt} if  and only if 
$$
c_j= \frac{1}{2 \Delta} \log \frac{W^{+,j}}{W^{-,j}}, \qquad\qquad\qquad j=1,..., p;
$$
in which case   $ \sigma_{\Delta}=0$.

(iii)  Let  $W^{+,j}\, W^{-,j} >0$ for all $j=1,..., p';$  and $W^{+,j}\, W^{-,j} =0$ for all $j=p'+1,..., p;$
and if
$$
c_j= \frac{1}{2 \Delta} \log \frac{W^{+,j}}{W^{-,j}}, \qquad\qquad\qquad j=1,..., p';
$$
then for any $\epsilon >0$
$$
 e^{ -\int_{0}^{\Delta}   \sigma_s \,  ds} \leq  \sum_{j=1}^{p'} 2\sqrt {W^{+,j}\;  W^{-,j}}+\epsilon,
$$
by an appropriate choice of $c_j$ for all $ j=p'+1,...,p.$
\end{thm}

{\it Proof.}  (i) It can be verified directly that
$$
\int_0^{\Delta} \sigma_s \, ds= -\log\[ \sum_{k=1}^m  w_0(k) e^{-\Delta y_k v(x_k)}\].
$$
Therefore,
$$
e^{-\int_0^\Delta \sigma_s \, ds}= \sum_{k=1}^m  w_0(k) e^{-\Delta y_k v(x_k)} =\sum_{j=1}^{p} W^{+,j}e^{-\Delta c_j}+ W^{-,j} e^{\Delta c_j}.
$$
Inequality \ref{ttt} follows from the inequality between arithmetic  and geometric means.

Parts {\it (ii)} and {\it (iii)} follow from this formula similar to the proof of Theorem \ref{pop}.
\qed

The theorem suggests the following procedure. We put $\Delta_p=1$ for all $p=0,1,2,....$  On each round of  the  boosting procedure,   we pick  $h_\gamma, \; \gamma \in \Gamma;$ such that the corresponding 
sum
$$
Z=\sum_{j=1}^{p} 2\sqrt {W^{+,j}\;  W^{-,j}},
$$
is minimal over the set of all weak classifiers. By adjusting the  values of $h_\gamma$  according to formulas of  Theorem 3.7  we minimize the penalty function
$\E$ on this round in an optimal way.

Let us comment on the  square roots which appear in the formula for $Z$. The set of all values of $h_{\gamma}$ is  a finite set $c_j, \; j=1,..., p.$ 
Therefore,  we have two special points of $p-1$ dimensional simplex  of probability measures
$$
p^+=\frac{1}{W^+} (W^{+,1},..., W^{+,p}),
$$
and
$$
p^-=\frac{1}{W^-} (W^{-,1},..., W^{-,p}).
$$
It is apparent that
$$
Z=\sum_{j=1}^{p} 2\sqrt {W^{+,j}\;  W^{-,j}}= 2\sqrt{W^+ W^-} BC(p^+, p^-).
$$
where $BC(p,q)$ is a Bhattacharyya coefficient, \cite{Bh}, the standard  measure a separability of classes in classification.

\subsection{The AdaBoost Flow as a Gradient System.  }
In this section we  write  equations \ref{evcl}--\ref{evme} in the gradient form  and obtain geometrical description of the algorithm.

A function  $D (v\, || \,w)$  where $v,w \in \W$    defined on a space $\W$, is called a divergence function, \cite{ANA}, 
when it satisfies the following conditions:

1) $D (v\, || \,w)  \geq 0.$

2) $D (v\, || \,w)  = 0$, when and only when $v = w$.

3) For small $dw$, Taylor expansion gives
$$
D (w + dw\, || \,w) \approx \frac{1}{2}  \sum g_{ij} dw(i) dw(j),
$$
where $g_{ij}=g_{ij}(w)$ is a positive-definite matrix. A divergence is not a distance because it is not necessarily symmetric with respect to $v$  and $w$, and it does
not satisfy the triangular inequality. 

The Kulback-Leibler divergence 
$$
KL(v\, || \,w) =\sum_{k=1}^{m}   v(k) \log \frac{v(k)}{w(k)}
$$
expanded near diagonal, see \cite{ANA},  produces the metric
$$
g_{ij} =\frac{\delta_i^j}{  w(i)}. 
$$

Using substitution $w_t(k)= r_t^2(k)$ and $\lambda_t(k) = y_k v(H_t,w_t)(x_k)$  equations \ref{evme} can be reduced to the form
$$
\frac{d}{d t}\;  r_t{(k)}= -   \frac{1}{2}\(\l_t(k)  -   \sigma_t\)  r_t(k), \qquad \quad  \qquad k=1,2, ..., m.   
$$
Moser,  \cite{M74},  noted that  on the surface of the sphere
$$
\sum_{p=1}^{m} r_t^2(p) =1,\qquad\qquad r_t(p)>0;  
$$
these equations can be put in the gradient form 
$$
\frac{d}{d t}\;  r_t{(k)}= -   \frac{\partial V}{\partial r_t{(k)} }, \qquad \quad  \qquad k=1,2, ..., m;
$$
with
$$
V=\frac{\sum_{p=1}^{m} \l_t(p) r_t^2(p)  }{4\sum_{p=1}^{m} r_t^2(p) }. 
$$
From this fact we have immediately 
\begin{lem} Equations \ref{evme} can be written in the gradient form
$$
\frac{d}{d t}\;  w_t(k) =- \nabla^k  V =-\sum_{j=1}^{m} g^{kj} \frac{\partial V}{\partial w(j)},  \qquad \quad  \qquad k=1,2, ..., m;
$$
with the  metric $g^{kj}$  defined as
$$
g^{kj}=\delta_k^j  w_t(k), 
$$
and the potential function 
$$
V_{h}(w)= \frac{\sum_{p=1}^{m} \l_t(p) w_t(p)  }{\sum_{p=1}^{m} w_t(p) }.  
$$

\end{lem}
In other words the Ada Boost flow on measures \ref{evme}   is a gradient flow produced by the function $V_{h}$ with respect to the  Kulback-Leibler metric. 
Now we can give an invariant geometric description of the AdaBoost algorithm.

Consider a foliation of  $R^m_+,  \; \W \in R^m_+,$ by  level sets of the function
$$
V_{h}(w)=\frac{\sum_{p=1}^{m} y_p h(x_p)  w(p)  }   {\sum_{p=1}^{m} w(p) }, \qquad\qquad \qquad w\in R^m_+. 
$$
Namely,  each leaf of the foliation is defined as
$$
\{w\in R^{m}_{+}  : V_h(w)=  c\},
$$
where $c$ is an arbitrary constant. 

Initial step of the algorithm. Pick $w_0$ and chose $h_{\gamma_0}$ such that 
$$
V_{h_{\gamma_0}}(w_0)  > 0
$$
takes maximal value. Namely,
$$
V_{h_{\gamma_0}}(w_0) =\max_{h_0  \in \H_0} V_{h_0}(w_0). 
$$
Run the AdaBoost flow \ref{evcl}--\ref{evme}  with this $h_{\gamma_0}$ until the  moment $t_0$ when the  flow reaches the leaf 
$\{w: V_{h_{\gamma_0}}(w)=0\}$.  Take $H_{t_0}=t_0 h_{\gamma_0}$ as a result of the initial  step.

Repeat the procedure with  $w_0$ replaced by $w_{t_0}$, {\it etc.}.

\subsection{General geometric approach to boosting. } 

The idea to define updates of the measure through solutions of a differential equation seems to be very  general. To define boosting algorithm one needs three ingredients (i)  the metrics; (ii) the potential function; (iii) the rule for  changing a potential function.  

Let us illustrate how this approach  produces Logitboost  algorithm.  We modify in  the  original AdaBoost algorithm the metrics and the potential function.  
Define    binary Kulback-Leibler divergence
$$
BKL(v\, || \,w) =\sum_{k=1}^{m}  \[ v(k) \log \frac{v(k)}{w(k)}  + (1-v(k)) \log \frac{1- v(k)}{1-w(k)}\]. 
$$
It produces the metrics
$$
g_{ij} =\delta_i^j\[ \frac{1}{  w(i)} + \frac{1}{1- w(i)}\].
$$
Taking,  
$$
V_h(w)=\sum_{p=1}^{m} y_p h(x_p)  w(p) 
$$
we obtain  logistic equations
$$
\frac{d}{d t}\;  w_t(k) =- \nabla^k  V_h =  -y_k h(x_k) (1-w_t(k))w_t(k). 
$$
The explicit formulas for a solution 
$$
w_t(k)= \frac{w_0(k)}{(1-w_0(k)) e^{y_k h(x_k) t}  + w_0(k)}
$$ 
produce standard formulas for updates of the measure on each step, see \cite{FS12}. 

As we see well known algorithms correspond the cases when differential equations have an explicit formulas for solution.
In the case of general metrics obtained from  the Bregman divergence an explicit formulas  for updates of the measure may not exist. In these case one can solve differential equations by some numerical procedure.

Now we want to present the SuperBoost, a different modification of the AdaBoost algorithm corresponding to  a new choice of control. 
The metric and the potential function are the same as for the AdaBoost. 
It is a greedy algorithm  which for each moment of time $t\geq 0$  chooses
a weak classifier $h$ with the largest $\sigma_t(h)$.

Initialization,
\begin{itemize}
\item  $H_{-1}=0.$
\item  $w_{0}(i)=\frac{1}{m},\; i=1,\ldots,m.$
\item Choose weak classificator $h_{\gamma_{0}}\in\mathcal{H}_{0}$ such that :
$\sigma_{w_0} (h_{\gamma_{0}}) =  \max_{h\in \H_0} \sigma_{w_0} (h).$
\end{itemize}

Updates are occurring on each infinitesimal step $t\rightarrow t+dt$
\begin{itemize}
\item Change classifier $h_{\gamma_t}=h$  on a new $h_{\gamma_{t+dt}}=h'$ if $$\sigma_t(h)= \sigma_t(h')\quad\text{and}
\quad \frac{d}{dt}\; \sigma_t(h)< \frac{d}{dt}\;  \sigma_t(h').$$
\item Update the measure: $w_{t+dt}(i)=\frac{1}{Z}\exp(-dty_{i}h_{\gamma_t}(x_{i}))w_t(i).$
\item Update the resulting classifier: $H_{t+dt}=H_{t}+dt \times  h_{\gamma_t}.$
\end{itemize}

\noi
It is obvious   that  SuperBoost algorithm  for each finite time interval $[0,T]$ updates  the  weak classifier  only a  finite number of times. 

Gradient dynamics on measures defined above is different from the restricted gradient minimization procedure for the cost function  proposed in \cite{MBBF, FR}. Their  procedure  is  connected to the choice of new weak classifier on each step of boosting or in our language to the specification of control.   

\subsection{Boosting and Perelman's ideas for the Ricci flow.}
In this section we present  the most unexpected  consequence  of  our approach.  We describe a striking similarity between the AdaBoost flow  and Perelman's ideas, \cite{Pr},  to control the Ricci  flow
\beq\label{RI}
\frac{d}{dt}\,  g_t= -2 Ric _{g_t}
\eeq
where $Ric _{g}  $   is the Ricci tensor of the metric $g$ and  $g \in \M$ space of metrics on a Riemannian manifold $M$. In our notations we follow \cite{AN}  and \cite{TO}. The equation describes some optimization procedure in the space $\M$.   Perelman  extends that phase space. Namely he defines 
Gibbsian type   measure $dw$ on $M$ as 
$$
dw= e^{-f} d V_g,
$$
where $d V_g$ is a volume element constructed from the metric $g$.  Apparently  for given  $g$   the measure  
$dw$ can be identified with the potential function $f$. Now the flow  is defined on the extended phase space $\M\times C^{\infty}$.   In order to control singularities of the Ricci flow $g_t,\; t \geq 0;$ Perelman  chooses the  potential  function $f$ in a special way   determined by  dynamics of the metric. 
The   system of coupled equations for the metric $g$ and potential function $f$ 
\beq\label{RICCI}
\frac{d}{dt}\,  g_t= -2( Ric _{g_t} +Hess _{g_t}f_t),  
\eeq
\beq\label{evmes}
\frac{d}{dt}\,  f_t= -R_{g_t} -\Delta f_t, 
\eeq
where $R_{g}$ is the scalar curvature of the metric $g$, leads to 
$$
\frac{d}{d t} (d w)=\frac{d}{d t} \( e^{-f_t} d V_{g_t}\)=0.
$$
The  flow  defined by \ref{RICCI}  is the original Ricci flow \ref{RI} up to a time dependent diffeomorphism.  

These equations are analog of the AdaBoost flow equations \ref{evcl} and \ref{evme}. To be precise equation \ref{evmes} is an  analog of \ref{evmepot}  which is an equivalent form of  \ref{evme}. Moreover, these equations are  similar termwise. The scalar curvature $R_{g_t}$ in  \ref{evmes} plays the role similar to that of  the margin $y_k v(H_t,w_t)(x_k)$ in \ref{evmepot}. The Laplacian $\Delta f_t$ in  \ref{evmes} is similar to the term  $\sigma_t$ in \ref{evmepot}. 

On the   extended  phase  space $\M\times C^{\infty}$  Perelman  defines the following functional 
$$
\F(g, f)= \int_{M}\( R_g+ |\bigtriangledown {f} |^2\)e^{-f} d V_g.
$$
Perelman calls the functional $\F(g, f)$   entropy for the Ricci flow. The functional  
$\F(g, f)$ increases along trajectories of the Ricci flow. Indeed, the formula 
$$
\frac{d}{d t} \F(g_t, f_t)= \int_{M}\langle - Ric_{g_t}- Hess_{g_t} f_t ,\frac{d g_t}{d t} \rangle e^{-f_t} d V_{g_t} , 
$$
together with  equations \ref{RICCI} and \ref{evmes} leads to
\beq\label{incr}
\frac{d}{d t} \F(g_t, f_t)= 2 \int_{M}| Ric_{g_t} + Hess_{g_t} f_t |^2 e^{-f_t} d V_{g_t}\geq 0.  
\eeq
The functional $\F(g, f)$ 
 is an analog of the Lyapunov function $\E(H,w)$.  
As we saw in section 3.2 the  functional  $\E(H,w)$ for the Ada Boost flow is closely connected to the ordinary Kullback-Leibler entropy. It steadily  decreases along trajectories of the AdaBoost flow.

It is time to take  stock of  these similarities. The   dictionary between two problems is below
\newline

\begin{tabular}{|c|c|}
\hline $TS$ training set & $M$ Riemannian manifold\tabularnewline
\hline $\mathcal{H}$ cone over the set of classifiers &
$\mathcal{M}$ space of Riemannian metrics \tabularnewline \hline
$\mathcal{H}\times W$ phase space of the  & $\mathcal{M}\times
C^{\infty}$ phase space of the \tabularnewline AdaBoost flow &
controlled Ricci flow\tabularnewline 
\hline $\frac{d}{d t}\; \lambda_t^k=    v^k(H_t,w_t)$  &  $\frac{d}{dt}\,  g_t= -2( Ric _{g_t} +Hess _{g_t}f_t)$\tabularnewline
\hline $\frac{d}{d t}\;  f_t{(k)}=    y_k v(H_t,w_t)(x_k)  -   \sigma_t $  & $\frac{d}{dt}\,  f_t= -R_{g_t} -\Delta f_t$ \tabularnewline
\hline $\E(H,w)$  & $\F(g, f)$ \tabularnewline
\hline $\frac{d}{ d t} \log \E(H_t,w_0) = - \sigma_t$  & $\frac{d}{d t} \F(g_t, f_t)\geq 0 $ \tabularnewline

\hline
\end{tabular}
\newline

\noi
Such coincidence is not accidental. 
We refer  to the introduction where it is explained that the AdaBoost and the Perelman's construction are different realizations of the same idea.


\vskip .1in
\noindent
A.L and S.M
\newline
Faculty  of Mathematics and  Mechanics
\newline
Moscow State University
\newline
Vorobjevy Gory
\newline
Moscow
\newline
Russia
\vskip 0.1in
\noindent
stepan.muzychka@gmail.com
\newline
alekslyk@yandex.ru

\vskip .1in
\noindent
K.V.
\newline
Department of Mathematics
\newline
Michigan State University
\newline
East Lansing, MI 48824
\newline
USA
\vskip 0.1in
\noindent
vaninsky@math.msu.edu

\end{document}